\documentclass[sigconf]{acmart}
\usepackage{amsthm}
\usepackage{amsmath}
 \newtheorem{defn}{Definition}
  
 \usepackage{multirow}
 \usepackage{xcolor}
\usepackage{tcolorbox}
 \usepackage{tikz}
\usetikzlibrary{fit}
\usepackage{colortbl}
 \usepackage{subfigure}
\usepackage{comment}
\usepackage{float}
\usepackage{algorithm}
 \usepackage{algorithmicx}
 \usepackage{algpseudocode}
 \usepackage{caption}
\usepackage{array}
 \usepackage{tabularx}
\usepackage{diagbox} 
\usepackage{colortbl} 
\usepackage{diagbox} 
\usepackage{makecell} 
\newcolumntype{C}[1]{>{\centering\arraybackslash}p{#1}}

\AtBeginDocument{%
  \providecommand\BibTeX{{%
    \normalfont B\kern-0.5em{\scshape i\kern-0.25em b}\kern-0.8em\TeX}}}
\setcopyright{acmlicensed}
\copyrightyear{2018}
\acmYear{2018}
\acmDOI{XXXXXXX.XXXXXXX}

\acmConference[Conference acronym 'KMS]{In Proceedings of International KMS
Conference }{June 03--05,
  2018}{Woodstock, NY}
\acmISBN{978-1-4503-XXXX-X/18/06}

\begin{document}

\title{Topology-guided Hypergraph Transformer Network: Unveiling
Structural Insights for Improved Representation}
\author{Khaled Mohammed Saifuddin}
\email{ksaifuddin1@student.gsu.edu}
\authornotemark[1]
\affiliation{%
  \institution{Georgia State University}
  \city{Atlanta}
  \state{Georgia}
  \country{USA}
}

\author{Mehmet Emin Aktas}
\affiliation{%
  \institution{Georgia State University}
   \city{Atlanta}
  \state{Georgia}
  \country{USA}}
\email{maktas@gsu.edu}

\author{Esra Akbas}
\affiliation{%
  \institution{Georgia State University}
   \city{Atlanta}
  \state{Georgia}
  \country{USA}}
\email{eakbas1@gsu.edu}

\renewcommand{\shortauthors}{Saifuddin et al.}

\newcommand{\sathg}{\texttt{THTN}}
\begin{abstract}


Hypergraphs, with their capacity to depict high-order relationships, have emerged as a significant extension of traditional graphs. Although graph neural networks (GNNs) perform remarkably well in graph representation learning, their extension to hypergraphs encounters challenges due to their intricate structures. 
 Furthermore, current hypergraph transformers, a special variant of GNN, utilize semantic feature-based attention, ignoring topological attributes of nodes and hyperedges. To address these challenges, we propose a \underline{T}opology-guided \underline{H}ypergraph \underline{T}ransformer \underline{N}etwork (\sathg). In this model, we first formulate a hypergraph from a graph while retaining its structural essence to learn higher-order relations within the graph. Then, we design a simple yet effective structural and spatial encoding module to incorporate the topological and spatial information of the nodes into their representation. Further, we present a structure-aware attention mechanism that discovers the important nodes and hyperedges from both semantic and structural viewpoints.  By leveraging these two modules, \sathg\ crafts an improved node representation, capturing both local and global topological expressions. Extensive experiments conducted on node classification tasks demonstrate that the performance of the proposed model consistently exceeds that of the existing approaches. 
\end{abstract}

\ccsdesc[500]{Computing methodologies~Machine learning algorithms}
\keywords{Graph Neural Network, Hypergraph Neural Network, Hypergraph Transformer}

\maketitle

\section{Introduction}
Inspired by the success of transformers in natural language processing~\cite{vaswani2017attention}, the transformer architecture is extended to handle graph-structured data, resulting in graph transformers \cite{ ying2021transformers,chen2022nagphormer,zhang2022hierarchical,wu2022nodeformer, zhu2023structural, li2022kpgt}. 
Along with structure inductive bias, they learn expressive node and graph representations via aggregating weighted semantic-structural information from neighboring nodes in the graph \cite{yun2019graph, zhao2021gophormer}. 
These models leverage the inherent pairwise relational information present in graphs to learn meaningful node, edge, and graph representations by capturing the dependencies and interactions among nodes. On the other hand, in real-world scenarios, relationships often extend beyond simple pairwise interactions, exhibiting more complex and higher-order dynamics. Furthermore, hidden higher-order relations between nodes in standard graphs might not be captured effectively with these models.

\textit{Hypergraphs} are increasingly used to model higher-order interactions in complex systems where entities are represented as nodes, and higher-order interactions among them are represented as hyperedges. 
Despite its strength, there are not many available real-world datasets represented as hypergraphs. Therefore, it is common to convert a standard graph to a hypergraph capturing higher-order interactions in it. The basic method in the literature for the conversion involves aggregating nodes with similar semantics, based on their attributes, into hyperedges \cite{feng2019hypergraph, chen2020hypergraph, jiang2019dynamic}. While this approach is effective in capturing semantic similarities of nodes, it may result in a loss of detailed structural information in the standard graph.

While Graph Neural Networks (GNNs) show great performance in graph representation learning,  the complexity of hypergraph structure poses a significant hurdle in adapting these models for efficiently learning hypergraph representation. In response, researchers have introduced various hypergraph representation learning techniques, including hypergraph neural networks \cite{feng2019hypergraph}, hypergraph convolution networks \cite{bai2021hypergraph}, and hypergraph attention networks \cite{chen2020hypergraph}. These methods aim to learn node and hyperedge embeddings by considering both the topology and attributes of hypergraphs, offering improved solutions for understanding complex relationships and patterns within such data. 
Furthermore, drawing on the success of transformers in graph data analysis, the hypergraph transformer model has been developed to extend this approach to hypergraphs~\cite{li2023hypergraph, xia2022self,zhang2022hegel, ding2023hyperformer}. However, current hypergraph transformer models mainly focus on attribute-based semantic features in the attention layer,  resulting in the loss of the structural attributes of nodes and hyperedges. 


To address these limitations, we propose a novel  \underline{T}opology-guided \underline{H}ypergraph \underline{T}ransformer \underline{N}etwork (\sathg) model. \sathg\ generates the representation of nodes via a new structure-aware attention mechanism that
effectively identifies the importance of nodes and hyperedges from
both semantic and structural perspectives. 
As the initial step, we create hypergraphs from standard graphs by considering nodes' higher-order relations that preserve the structure information in the graph.
 Then, we define our \sathg\ model on the constructed hypergraph. In the input layer of \sathg\, we introduce a learnable \textit{structure encoding} scheme comprising local structure encoding, centrality encoding, and uniqueness encoding to capture important local and global structural information. Furthermore, we utilize a learnable hypergraph Laplacian eigenvector in the input layer as node positional information. This Laplacian eigenvector-based position encoder enables the encoding of distance-aware spatial information within the hypergraph. 

Moreover, while a hyperedge consists of multiple nodes, each node possesses varying degrees of structural and semantic importance for that hyperedge. Similarly, a node may belong to multiple hyperedges, but not all hyperedges hold equal structural and semantic importance for that node. To capture individual importance from both structural and semantic perspectives, we introduce a \textit{structure-aware attention} mechanism that consists of two layers. i) Local Structure-Aware Node-to-Hyperedge Attention layer aggregates node representations into a hyperedge representation via emphasizing structurally and semantically significant nodes. To define the structural importance of a node for a hyperedge, we use two measures, node-local clustering coefficient, and node coreness. These measures calculate how densely connected the node's local neighborhood is.
ii) Global Structure-Aware Hyperedge-to-Node Attention layer aggregates hyperedge representations into a node representation via highlighting structurally and semantically important hyperedges. To define the structural importance of a hyperedge for a node, we introduce two new measures, the hyperedge density score and the hyperedge clustering coefficient.  
Contributions of our work are summarized as follows:

\begin{itemize}
\item \textbf {Hypergraph Construction:} Unlike existing approaches that construct hypergraphs from input graphs by grouping nodes with semantically similar attributes into a hyperedge 
resulting in a loss of network structure information, we construct a hypergraph by discerning higher-order connections between nodes 
ensuring the preservation of node structural integrity.
\item \textbf{Node Feature Enrichment via Structural and Spatial Encoding:} 
\sathg\ introduces three different structure encoding schemes (local and global) along with a position (spatial) encoder, enhancing the model's ability to capture important structural and spatial information of nodes. These encoding schemes are incorporated alongside the input node features, allowing the model to effectively leverage both the initial features of nodes and the discovered structural-spatial patterns.
\item \textbf{Structure-Aware Attention: } We introduce a structure-aware attention mechanism in addition to the semantic feature-based vanilla attention utilized in existing hypergraph transformers. Our attention mechanism in \sathg\ incorporates both attribute-based semantic features and structural information that enables the identification of crucial nodes for a hyperedge and important hyperedges for a node from both points of view. We further introduce four different measures to discover the structurally important nodes and hyperedges.
\end{itemize} 
 

\section{Background}\label{sec:background}
In this section, we first outline the basics of transformer and hypergraph. Then, we review the literature on graph and hypergraph neural networks, including their transformer variants, emphasizing key contributions and findings in these areas. 
\subsection{Preliminaries}
\subsubsection{Transformer}\label{transformer}
The transformer is a neural network architecture that exploits an attention mechanism to capture local and global dependencies in the input data. The main components of the model are a multi-head self-attention (MHA) module and a position-wise feed-forward network (FFN). The MHA module computes attention weights by projecting the input into a query ($Q$), key ($K$), and value ($V$) vectors. If $Z \in \mathbb{R}^{n\times d}$ is a $d$ dimensional input feature where $Z=[z_1,z_2,....,z_n]^T$, the MHA first projects it to Q, K, and V using three learnable weight matrices $W^Q \in \mathbb{R}^{d \times d_Q}$, $W^K\in \mathbb{R}^{d \times d_K}$, and $W^V\in \mathbb{R}^{d \times d_V}$ as
\begin{equation}
    \begin{aligned}
        Q = ZW^Q, K = ZW^K, \text{ and } V = ZW^V.
    \end{aligned}
\end{equation}
Then in each head $h\in \{1,2,3,...,H\}$, the attention mechanism is applied to the corresponding ($Q_h,K_h,V_h$) as

\begin{equation}\label{eq2}
\begin{aligned}
    \Delta_h=(Q_h K_h^T)/\sqrt{d_K},
    \end{aligned}
\end{equation}
\begin{equation}
\begin{aligned}
    output_h=\text{softmax}(\Delta_h)V_h.
    \end{aligned}
\end{equation}
Here, $\Delta$ represents the attention map and the dimensions of $d_Q=d_K=d_V=d$. The outputs from different heads are concatenated and transformed to get the MHA output, which is further fed into a position-wise FFN layer. Residual connections and layer normalization are used to stabilize and normalize the outputs.

\subsubsection {Hypergraph}
A hypergraph allows us to represent more flexible relationships between entities compared to traditional graphs. It consists of degree-free hyperedges, which connect any number of nodes. Below is a formal definition of a hypergraph.

\begin{defn}
A hypergraph is defined as ${\mathcal{G}} = (\mathcal{V}, \mathcal{E})$ where $\mathcal{V} =\{v_{1},...,v_{m}\}$ is the set of nodes and $\mathcal{E} = \{e_{1},...,e_{n}\}$ is the set of hyperedges. 
Similar to the adjacency matrix of a standard graph, a hypergraph can be denoted by an incidence matrix $A^{m \times n}$ where $m$ is the number of nodes, $n$ is the number of hyperedges and $A_{ij}=1$ if $v_i \in e_j$, otherwise 0.
\end{defn}
\begin{figure*}[t!]
  \centering
  \includegraphics[width=\textwidth, height=9cm]{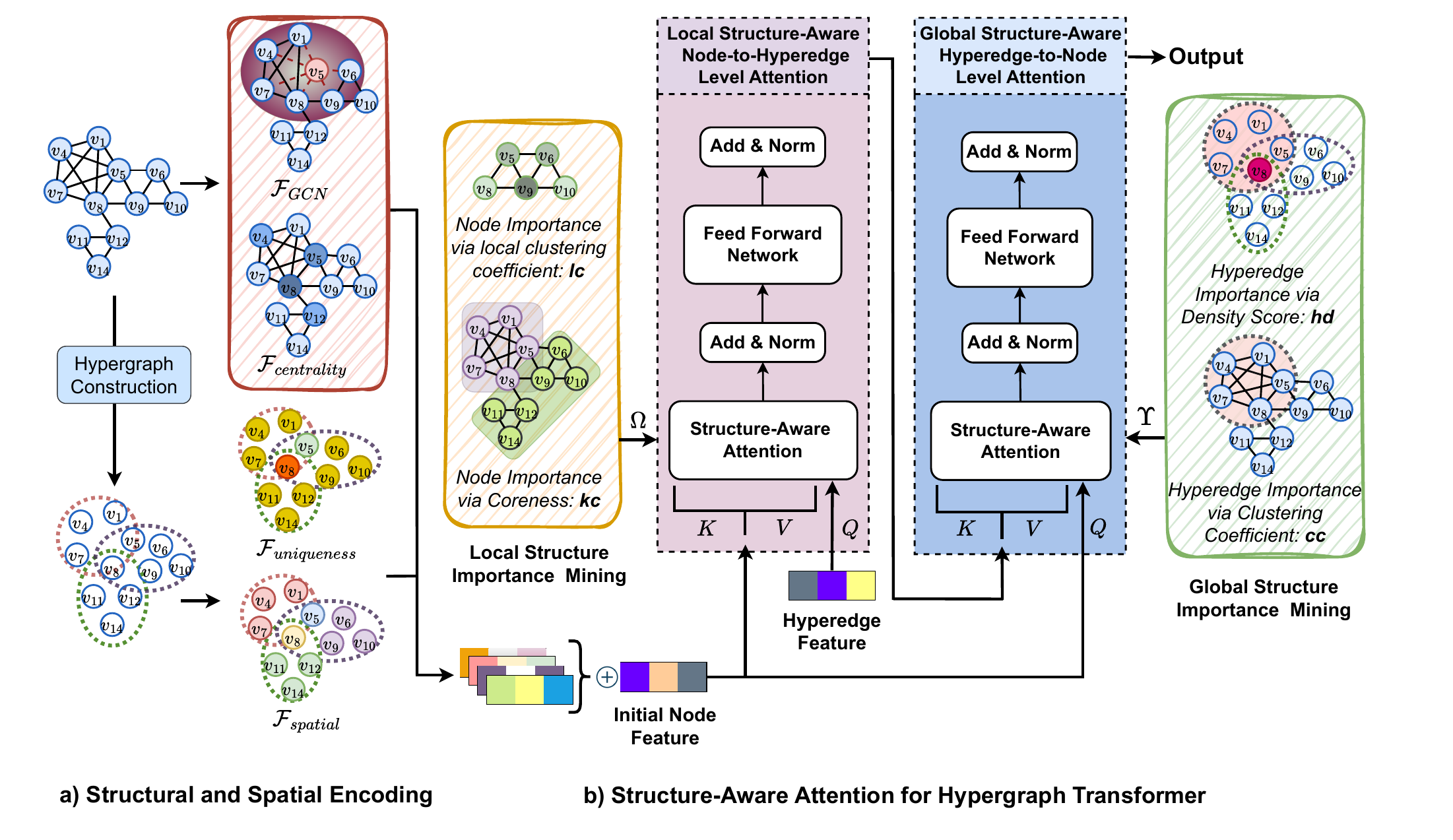}
  \caption{Topology-guided Hypergraph Transformer Network, \sathg, consists of a) \textit{Structural and Spatial Encoding} that enrich initial node representation via learnable structural and spatial node features and  b) \textit{Structure-Aware Attention} that enables the integration of structural importance  of
nodes for a hyperedge and hyperedges for a node into regular attribute-based semantic attention to derive the ultimate node representations.}
  \label{fig:thtn}
\end{figure*}

\subsection{Related Work} \label{sec:works}
Advancements in graph neural networks (GNNs) have prompted their extension to hypergraphs, yielding hypergraph neural networks that address unique challenges in hypergraph-structured data. Various models have been proposed in recent years, such as SubHypergraph inductive neural networks (SHINE) \cite{luo2022shine}, hypergraph attention networks (HANs) \cite{chen2020hypergraph}, Hyper-SAGNN model \cite{zhang2020hypersagnn}.
The Hypergraph Neural Networks (HGNN) \cite{feng2019hypergraph} extend GNNs to hypergraphs, enabling efficient handling of complex data correlations via hyperedge spectral convolution. Yet, HGNN overlooks the varying importance of node-hyperedge connections. HAN \cite{chen2020hypergraph} resolves this by adapting GATs' \cite{velivckovic2017graph} attention mechanism for hypergraphs, allowing for adaptive learning of node and hyperedge significance based on their contributions to the structure.


\citeauthor{vaswani2017attention} introduce the transformer model in \cite{vaswani2017attention}, emphasizing attention mechanisms' ability to capture long-range dependencies in sequential data. This inspired the development of graph transformers, extending attention to handle graph-structured data. \citeauthor{dwivedi2020generalization} pioneered in this domain by proposing a graph transformer \cite{dwivedi2020generalization} that incorporates graph connectivity as an inductive bias, ensuring nodes attend only to their neighbors. This model utilizes the graph Laplacian eigenvectors for node positional encoding. Gophormer \cite{zhao2021gophormer} takes a different approach, extracting ego-graphs as transformer inputs instead of processing the entire graph. Challenges in assimilating structural data into the transformer model are addressed through the proposal of a proximity-enhanced attention mechanism \cite{zhao2021gophormer} and by incorporating centrality, spatial, and edge encoding techniques \cite{ying2021transformers}. 


While transformers for standard graphs have seen considerable progress, a gap remains in developing transformers specifically for hypergraphs. Recent research in hypergraph transformers \cite{li2023hypergraph, liu2023meta, xia2022self,zhang-etal-2022-hegel, ding2023hyperformer} shows promise but tends to focus predominantly on the attribute-based semantic features of nodes and hyperedges within the attention module, while often overlooking crucial structural information. Notably, models in \cite{li2023hypergraph, liu2023meta} are limited to meta-path-guided heterogeneous hypergraphs, further narrowing their scope. This study addresses this gap by introducing a topology-guided transformer model for hypergraphs that consists of a structural and spatial encoding module and a structure-aware attention mechanism that considers both the semantic and structural importance of nodes and hyperedges. By incorporating both aspects, the proposed model provides a more comprehensive representation and understanding of the underlying data.

\section{Methodology}
\label{sec:method}
In this section, we outline components of our \sathg\ model. As the first step of \sathg, we present the structure and spatial encoding module as the initial node feature enrichment process. Next, we propose a structure-aware attention module that integrates both the local and global structural information of nodes and hyperedges into the regular semantic feature-based attention network as an inductive bias. 

In Section~\ref{sec:hycons}, we explain how we create our structure-based hypergraph from a standard graph. 
Unlike existing works that construct hypergraphs by grouping semantically similar nodes into hyperedges~\cite{feng2019hypergraph, chen2020hypergraph, jiang2019dynamic}, which may result in a loss of critical structural information, we develop a hypergraph from an input graph by identifying subgraphs considering higher-order structural relations of nodes and representing these subgraphs as hyperedges. 
We can consider different measures to define higher-order structures such as communities, cliques, and motifs. Details about hypergraph construction are presented in Section~\ref{sec:hycons}. We present the system architecture of our \sathg\ in Figure \ref{fig:thtn}. 
\vspace{-2mm}
\subsection{Node Feature Enrichment via Structural and Spatial Encoding}\label{structural_and_spatial}

In graph data, a node's significance extends beyond its individual attributes. Its position, connections, and unique characteristics within the entire graph determine its relevance. Traditional feature representation methods often miss this nuanced information, as two nodes might share attributes but differ in roles due to their connections and positions. Similarly, a uniquely attributed node might still be peripheral in the overall structure. Moreover, while we create hypergraphs from graphs, we lose the nodes' local connection information. Thus, in order to incorporate the important structural and spatial information of the nodes into their representations via a transformer, we introduce four different encoding methods from both graph and hypergraph perspectives: (1) local structure encoding $lse$, (2) centrality encoding $ce$, (3) uniqueness encoding $ue$ and (4) position encoding $pe$. We combine all these encodings and integrate them into initial (i.e., $0^{th}$ layer) node features of a node $v_i$, $x_i^{0}$, as follows
\begin{equation}
    \begin{aligned}
        x_i=x_i^{0}+lse_i+ce_i+ue_i+pe_i.
    \end{aligned}
\end{equation}
By integrating these encodings with the original input features, the transformer can better learn node representations, encoding both structural-spatial and semantic context. The subsequent sections will delve deeper into each encoding, elucidating their formulation and contributions.

\subsubsection{Local Structure Encoding} As mentioned earlier, we create a hypergraph from a given graph by representing subgraphs as hyperedges. However, while creating a hyperedge from a subgraph, we lose the local connection information between nodes, which might be crucial for the hyperedge representation and thus the hypergraph. To preserve the local connection information, we apply graph convolutional networks (GCNs) to the input graph. GCNs adeptly capture and retain local connection information by propagating information across node neighborhoods, thus discerning each node's local structure encoding ($lse$). This encoded information is combined with the input node features.
\begin{equation}
\begin{aligned}
\label{eqn_lse_encoding}
lse_i = \mathcal{F}_{GCN}\left( \mathcal{N}(i); \theta \right),
\end{aligned}
\end{equation}
where $lse_i$ denotes the local structure encoding for node $v_i$, calculated using a GCN function, represented as $\mathcal{F}_{GCN}$. This function is applied to the node's neighborhood, $\mathcal{N}(i)$, and is parametrized by $\theta$, a set of learnable parameters, to adaptively capture the localized information around node $v_i$.

\subsubsection{Centrality Encoding} In line with our hypergraph construction strategy, we employ closeness centrality to discern the significance of nodes within the graph structure. Nodes with high closeness centrality are integral for efficient information propagation due to their shorter distances to other nodes. In \sathg, we leverage closeness centrality as an additional signal to enhance the input node features. To be specific, we develop a learnable centrality encoding function $\mathcal{F}_{centrality}$; given the closeness centrality scores as input; it generates a centrality embedding vector for each node. By integrating the original nodes' features with these embedding vectors,  the transformer is better equipped to grasp the nodes' roles and influence within the graph. Centrality encoding $ce$ can be formulated as
\begin{equation}
\begin{aligned}
\label{eqn_ce_sophisticated}
ce = \mathcal{F}_{\text{centrality}}(c; \psi),
\end{aligned}
\end{equation}
where $c$ is a vector of centrality scores of all the nodes, and $\psi$ is a learnable parameter.

\subsubsection{Uniqueness Encoding} If a node appears in multiple hyperedges, it may not possess a distinct identity within any specific hyperedge, and as a result, its significance within those hyperedges could be diminished. To encode the importance of nodes based on their appearance in different hyperedges, we introduce a uniqueness score $\mathcal{u}$, and for node $v_i$, defined as
\begin{equation}
\begin{aligned}
    \mathcal{u}_i=1-\frac{C_A(i)}{{C_T}},
\end{aligned}
\end{equation}
where $C_A(i)$ is the number of hyperedges that node $v_i$ appears, $C_T$ is the total number of hyperedges in the hypergraph. The more a node appears in different hyperedges, the less uniqueness the score is.
We store the uniqueness scores of all the nodes into a vector $u$. Given the $u$ as an input, we exploit a learnable function $\mathcal{F}_{uniqueness}$ that assigns an embedding vector to each node according to its uniqueness score, which is further added with the input node features. Uniqueness encoding $ue$ is defined as
\begin{equation}
\begin{aligned}
\label{eqn_ue_detailed}
ue = \mathcal{F}_{\text{uniqueness}}\left(u; \zeta \right),
\end{aligned}
\end{equation}
where $u$ is a vector of uniqueness scores of all the nodes, and $\zeta$ is a learnable parameter.

\subsubsection{Position Encoding}
Position encoding ($pe$) plays a crucial role in the transformer architecture for capturing sequential information using different position functions (e.g., sine, cosine). However, when dealing with arbitrary graphs, the direct application of positional encoding becomes challenging due to the absence of a clear positional notion. To overcome this limitation, researchers have introduced the use of eigenvectors derived from the graph Laplacian as a graph-specific alternative to sine and cosine functions \cite{dwivedi2020generalization, kreuzer2021rethinking}.

We adopt the method in \cite{rodriguez2009laplacian} to calculate the hypergraph Laplacian eigenvector $ev$. Then, we apply a learnable function, $\mathcal{F}_{\text{spatial}}$, which, when provided with the $ev$, produces an embedding for each node, serving as the node positional encoding. This captures the nodes' structural information, considers their relationships, and incorporates the connectivity patterns within the hypergraph. It also encodes distance-aware (spatial) information where nearby nodes or nodes connected by hyperedges share similar positional features. We incorporate this positional information with the node input features. Additionally, we address eigenvector multiplicity by randomly flipping the sign during training. This regularization technique prevents the model from overfitting to specific sign patterns and promotes the learning of more robust and generalized representations. Position encoding $pe$ can be expressed as

\begin{equation}
\begin{aligned}
\label{eqn_pe_spatial_aware}
pe = \mathcal{F}_{\text{{spatial}}}(ev; \phi),
\end{aligned}
\end{equation}
where $ev$ is hypergraph Laplacian eigenvectors and $\phi$ is a learnable parameter.

\subsection{Structure-Aware Attention}  
In this section, we describe our structure-aware attention mechanism capturing high-order relationships between nodes and hyperedges. This model incorporates novel two-level attention, encompassing both node-to-hyperedge and hyperedge-to-node information propagation via considering their semantic and structural importance for each other, facilitating the integration of local and global contexts into node representation learning.

The first level, node-to-hyperedge, aggregates node information to create hyperedge representations via local structure-aware attention. Conversely, the second level, hyperedge-to-node, gathers hyperedge representations to generate the final node representation via global structure-aware attention. As shown in Equation \ref{eq2}, the attention network in traditional transformer architecture calculates attention scores between different nodes and hyperedges by evaluating the attribute-based semantic similarity whereas the vital structural information is ignored. To address this, at each attention level, we introduce different structural inductive biases that allow the model to capture both semantic and vital structural information simultaneously.

\subsubsection{Local Structure-Aware Node-to-Hyperedge Level Attention.} Hyperedge is degree-free and consists of an arbitrary number of nodes. However, the contribution of nodes and their importance in hyperedge construction could be different. To identify the important nodes for a given hyperedge, we leverage an attention mechanism that aggregates node representations and assigns higher weights to crucial ones. In addition to a semantic perspective of regular attention, we inherit the local structure of subgraphs as hyperedges to extract the structural significance of nodes within hyperedges. Since we represent each subgraph as a hyperedge, the structurally significant nodes for a subgraph will also be important for that corresponding hyperedge. To determine these nodes, we calculate the structural importance of nodes for that subgraph (hyperedge). Then, we combine nodes' structural importance with the attention map as a structure inductive bias. With the modified attention mechanism, the $l$-th layer representation $q_{j}^l$ of a hyperedge $e_j$ is defined as
\begin{equation}
\begin{aligned}
\label{eqn_conditional_q}
q_{j}^{l} = \alpha \left( \sum_{v_{i}} [\Gamma_{ji} W_{1} p_{i}^{l-1} \mid v_{i} \in e_{j}] \right),
\end{aligned}
\end{equation}
where $\alpha$ is a nonlinear activation function, $W_1$ is a trainable weight matrix, and $\Gamma_{ji}$ is the attention coefficient of node $v_i$ in the hyperedge $e_{j}$.

\begin{equation}
\begin{aligned}
\label{eqn_conditional_x}
\Gamma_{ji} =  \frac{\exp({r_{ji})}}{\sum_{v_{k}} [\exp({r_{jk}}) \mid v_{k} \in e_{j}]},
\end{aligned}
\end{equation}
and
\begin{equation}
\begin{aligned}
\label{eqn_conditional_v}
{r_{ji}} =  \frac{\beta (W_{2} p_{i}^{l-1} * W_{3} q_{j}^{l-1})}{\sqrt{d_K}} + \Omega_{ji},
\end{aligned}
\end{equation}
where $\beta$ is a LeakyReLU activation function, $v_{k}$ is the node that belongs to hyperedge $e_{j}$, $W_2$, $W_3$ are the trainable weight matrices, $*$ is the element-wise multiplication, $\Omega_{ji}$ is the total structural importance of node $v_i$ for hyperedge $e_j$. We first start by presenting how we calculate nodes' structural importance, namely the local clustering coefficient ($lc$) and $k$-core ($kc$) values. We will eventually add these values to get the total local structural importance of a node for a hyperedge as
\begin{equation}
    \Omega_{ji}=lc_{ji}+kc_{ji}.
\end{equation}

\textit{i. Node Importance via Local Clustering Coefficient. }
Understanding the structural significance of individual nodes within a network is crucial for various applications such as information dissemination and influence assessment. One effective way to quantify a node's structural importance is by examining its local clustering coefficient. For our subgraphs, it can be defined as follows.

\begin{defn}[Node local clustering coefficient]
The local clustering coefficient ($lc$) of a node in a subgraph is the ratio of the number of connections between the node's neighboring nodes (within the subgraph) to the total number of possible connections between them (within the subgraph), i.e., 
\[
lc_{ji} = \frac{I_{ji}}{\frac{g_{ji}(g_{ji} - 1)}{2}}.
\]
In this formula, \( I_{ji} \) represents the number of connections existing among the neighbors of the node \( v_i \) within the subgraph that pertains to hyperedge \( e_j \). Conversely, the term \( \frac{g_{ji}(g_{ji} - 1)}{2} \) computes the maximum possible number of connections among all neighbors of \( v_i \), where \( g_{ji} \) indicates the degree of the node \( v_i \) in the subgraph associated with hyperedge \( e_j \).

\end{defn}

The $lc$ of nodes measures the density of connections among their neighbors, indicating the presence of tightly-knit clusters. Nodes with high $lc$ are important as they are likely to facilitate information flow, influence spreading, and contribute to network resilience. 





\textit{ii. Node Importance via Coreness. } The concept of $k$-core decomposition is rooted in the identification of a graph's maximal subgraphs, where each node is connected to at least $k$ other nodes within the network. It assigns a $k$-core ($kc$) value to each node, representing the highest level of connectivity it shares with its neighbors. Consequently, nodes with high $kc$ values are considered more central and pivotal within the network. The process begins by assigning a $k$-core value to each node of the subgraph, quantifying its connectivity level within the network. The decomposition process then systematically prunes nodes with lower connectivity, starting from those with the least connections and progressively moving towards nodes with higher degrees of connectivity. This pruning process is dynamic, with the connectivity degrees of neighboring nodes adjusted accordingly. Nodes persisting in the highest $k$-core constitute the network's core structure, showcasing robust connections.


\subsubsection{Global Structure-Aware Hyperedge-to-Node Level Attention. }

In hypergraphs, a node can be part of several hyperedges, yet not all hyperedges hold equal significance for a node. This necessitates an attention mechanism to emphasize the key hyperedges associated with a specific node. Moreover, employing vanilla attention as defined in Equation \ref{eq2} will only capture the importance of hyperedges from a semantic point of view, ignoring structural importance. To address this issue, we dive into the structural properties of hyperedges, and the corresponding subgraphs and calculate the structural importance of the subgraphs (hyperedges) for each node.

After determining the structural importance of hyperedges, we combine them with the attention map of the hyperedge-to-node level attention network as a structure inductive bias. With the modified attention mechanism, the $l$-th layer representation $p_{i}^l$ of node $v_i$ is defined as
\begin{equation}
\begin{aligned}
\label{eqn_conditional_p}
p_{i}^{l} = \alpha \left( \sum_{e_{j}} [\Lambda_{ij} W_{4} q_{j}^{l} \mid e_{j} \in E_{i}] \right),
\end{aligned}
\end{equation}
where $\alpha$ is a nonlinear activation function, $W_4$ is a trainable weight matrix, $E_i$ is the set of hyperedges connected to node $v_i$, and $\Lambda_{ij}$ is the attention coefficient of hyperedge $e_j$ on node $v_{i}$.



\begin{equation}
\begin{aligned}
\label{eqn_conditional_y}
\Lambda_{ij} = \frac{\exp({t_{ij})}}{\sum_{e_{k}} [\exp({t_{ik}}) \mid e_{k} \in E_{i}]},
\end{aligned}
\end{equation}
and
\begin{equation}
\begin{aligned}
\label{eqn_conditional_e}
{t_{ij}} =  \frac{\beta (W_{5} q_{j}^{l} * W_{6} p_{i}^{l-1})}{\sqrt{d_K}} + \Upsilon_{ij},
\end{aligned}
\end{equation}
where $W_5$ and $W_6$ are the trainable weight matrices, and $\Upsilon_{ij}$ is the total structural importance of hyperedge $e_j$ for node $v_i$. To calculate the structural significance of a hyperedge for its member nodes, we introduce two new measures: hyperedge density ($hd$) and hyperedge clustering coefficient ($cc$). By computing these measures, we can determine the relative importance of hyperedges in terms of their structural impact on the connected nodes. Calculated measures are simply added to get the total global structural importance of a hyperedge for a node as
\begin{equation}
\begin{aligned}
    \Upsilon_{ij}=hd_{ij}+cc_{ij}.
    \end{aligned}
\end{equation}

\begin{algorithm}[t]
\begin{algorithmic}[1]
\State \textbf{Input:} Subgraph $\mathcal{H}=(\mathcal{V_H},\mathcal{E_H})$
\State \textbf{Output:} Clustering Coefficients $cc$ of $\mathcal{H}$
\State $nodes \leftarrow \text{list}(\mathcal{V_H})$
\State $triangle\_count \leftarrow 0$
\For{$(idx, u) \in \text{enumerate}(nodes)$}
    \For{$v \in nodes[idx + 1:]$}
        \If{$\mathcal{H}\text{.has\_edge}(u, v)$}
            \State $com\_neig \gets \text{list(common\_neighbors}(\mathcal{H}, u, v))$
            \State $triangle\_count \gets triangle\_count + \text{len}(com\_neig)$
        \EndIf
    \EndFor
\EndFor
\State $triangle\_count \leftarrow triangle\_count/3$ \Comment{Each triangle is counted three times}
\State $m_{\mathcal{H}} \leftarrow |\mathcal{V_H}|$
\State $total\_possible\_triangles \leftarrow \frac{m_{\mathcal{H}}(m_{\mathcal{H}} - 1)(m_{\mathcal{H}} - 2)}{6}$
\If{$total\_possible\_triangles > 0$}
    \State $clustering\_coefficient \leftarrow \frac{triangle\_count}{total\_possible\_triangles}$
\Else
    \State $clustering\_coefficient \leftarrow 0.0$
\EndIf \\
\Return $cc \leftarrow clustering\_coefficient$
\end{algorithmic}
\caption{Hyperedge Clustering Coefficient Computation}\label{alg:cc} 
\end{algorithm}

\textit{i. Hyperedge Importance via Density Score. }
The hyperedge density ($hd$) is a measure that quantifies the proportion of nodes that a hyperedge $e_j$ consists of compared to the total number of nodes in the hypergraph. If a hyperedge $e_j$ consists of $m_{e_j}$ number of nodes and $m$ is the total number of nodes in the hypergraph, then $hd$ can be expressed as $\frac{m_{e_j}}{m}$. A higher $hd$ indicates stronger interconnectivity among the nodes within the hyperedge, implying a more cohesive and significant grouping. Hyperedges consisting of a larger number of nodes are deemed more significant for a given node compared to hyperedges with fewer nodes.

\textit{ii. Hyperedge Importance via Clustering Coefficient. }
In hypergraphs, hyperedges are crucial for representing complex relationships and patterns. To evaluate the structural importance of these hyperedges, we introduce the hyperedge clustering coefficient.
\begin{defn}[Hyperedge clustering coefficient]
The hyperedge clustering coefficient ($cc$) is defined from its subgraph, which is the ratio of the number of existing triangles between all pairs of nodes to the total number of possible triangles in that subgraph.
\end{defn}
The $cc$ provides insights into the cohesive structure of a subgraph by quantifying the extent of interconnectivity among its nodes. A higher $cc$ signifies that the nodes within a subgraph are densely interconnected, indicating a strong level of cohesion. A subgraph (hyperedge) with a larger $cc$ score is considered important for a node. Algorithm~\ref{alg:cc} outlines the steps to compute the hyperedge clustering coefficient from the subgraph.

Like the transformer paper \cite{vaswani2017attention}, the output of each attention layer is fed into a position-wise FFN layer. Moreover, residual connections and layer normalization are used to stabilize and normalize the output. \sathg\ generates the node representations by utilizing these two levels of attention. Finally, the output of \sathg\ is linearly projected with a shared trainable weight matrix $W_S$  to generate an $S$ dimensional output for each node as $O = pW_S$, where $S$ is the number of classes, $p$ is the output of the \sathg. We train our entire model using a cross-entropy loss function.

\subsection{Hypergraph Construction}
\label{sec:hycons}
Existing models for creating hypergraphs from standard graphs typically utilize node attributes and methods like $k$-means/$k$-NN to group semantically similar nodes into a single hyperedge~\cite{feng2019hypergraph, chen2020hypergraph, jiang2019dynamic}. That may cause a loss of structural insight in the graph, which is vital for effective node representation learning in a hypergraph setting. To address this challenge,
We develop a hypergraph from an input graph by identifying subgraphs considering higher-order structural relations of nodes and representing these subgraphs as hyperedges. We can consider different measures to
define higher-order structures such as communities, cliques, and
motifs.

In this study, we use communities to represent higher-order structural relations of nodes. Communities are densely connected subgraphs capturing group interaction beyond the pairwise interaction. While representing each community as a hyperedge, there should be overlaps between communities to make the hypergraph connected. To identify these communities, we evaluate various overlapping community detection methods and eventually adopt the algorithm described in \cite{chen2010detecting}, which exhibits better performance in our experiments. This algorithm utilizes edge attributes as weights; in cases where the input graph does not provide these attributes, we assign a uniform weight of 1 to each edge. Each detected community is then represented as a hyperedge in the hypergraph, with the community members serving as nodes within these hyperedges.

\begin{figure}[t!]
  \centering
  \includegraphics[width=.48\textwidth]{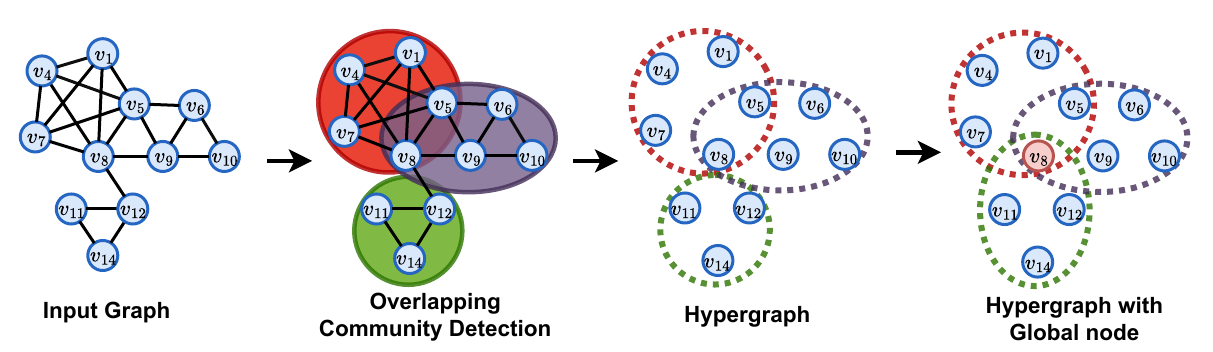}
  \caption{Hypergraph Construction: Each community is represented as a hyperedge.}
  \label{fig:hyg}\vspace{-3mm}
\end{figure}

While employing an overlapping community detection algorithm, it's notable that not all communities will overlap, occasionally resulting in isolated hyperedges that lack connections to others. This isolation can hinder the flow of information within the hypergraph. To mitigate this and enhance connectivity between hyperedges, we strategically incorporate global nodes chosen based on their high closeness centrality scores from the input graph. Closeness centrality, measuring the average shortest distance of a node to all other nodes, helps identify nodes that can rapidly disseminate information across the network. By linking these centrally located global nodes to hyperedges, we significantly boost the interconnectivity and information exchange across the hypergraph, ensuring a more cohesive and efficient network structure. Figure \ref{fig:hyg} provides a detailed visual representation of the hypergraph construction process.

\subsection{Complexity Analysis}
Given the $f$ dimensional initial feature of a node, \sathg\ exploits a two-level attention network and generates ${f}^{'}$ dimensional embedding vector for the node. Thus, the time complexity of \sathg\ can be expressed in terms of the cumulative complexity of the attention networks along with the structure encoding involved in each attention module. 

To formulate the total time complexity for the \textit{Local Structure-Aware Node-to-Hyperedge Level Attention}, we divide it into two parts. The first part involves calculating the time complexity of the node-to-hyperedge level attention, which can be formulated as $O(|\mathcal{V}|ff^{'}+|\mathcal{E}|d_{e}f^{'})$, wherein $|\mathcal{V}|$ represents the number of nodes, $|\mathcal{E}|$ represents the number of hyperedges, and $d_e$ denotes the average size of each hyperedge, typically a minimal value. In the second part, we calculate the time complexity of the local clustering coefficient and $k$-core. The local clustering coefficient exhibits a time complexity of $O(|\mathcal{E}|d_e)$. Analogously, the time complexity of the $k$-core decomposition can be expressed as $O(|\mathcal{E}|d_e)$. So, the overall time complexity involved in \textit{Local Structure-Aware Node-to-Hyperedge Level Attention} is 
 $O(|\mathcal{V}|ff^{'}+|\mathcal{E}|d_{e}f^{'})$+2$O(|\mathcal{E}|d_e)$.

Similarly, if $d_n$ represents the average node degree (i.e., the average number of hyperedges to which a node belongs), the time complexity for the hyperedge-to-node level attention can be expressed as $O(|\mathcal{E}|f{f}^{'}+|\mathcal{V}|d_n{f}^{'})$. The cumulative time complexity required to calculate $hd$ is demonstrable $O(|\mathcal{V}|d_n)$. Consequently, the overarching time complexity incurred for determining the hyperedge clustering coefficient is calculated to be $O(|\mathcal{E}|(d_{e}c_{n})^{3/2})$, where $c_n$ is the average number of common neighbors between any two nodes in a hyperedge. So, the overall time complexity involved in Global Structure-Aware Hyperedge-to-Node Level Attention is $O(|\mathcal{E}|f{f}^{'}+|\mathcal{V}|d_n{f}^{'})$+$O(|\mathcal{V}|d_n)$+$O(|\mathcal{E}|(d_{e}c_{n})^{3/2})$.

\section{Experiment} \label{sec:experiment}
\subsection{Experimental Setup}
We evaluate \sathg\ on six distinct datasets including citation network and social network datasets~\cite{Fey/Lenssen/2019}. Detailed statistics for these datasets are provided in Table \ref{dataset_statistics}.  The community detection algorithm follows \cite{chen2010detecting} with default settings. We also present constructed hypergraph statistics of these datasets in Table \ref{dataset_statistics}. The number of global nodes for the Cora and the Citeseer, PubMed, DBLP, LastFMAsia, and Wiki datasets is set to 3, 1, 4, 5, 4, and 4 respectively.  Our experiments follow a standard split of 50\% for training, 25\% for validation, and 25\% for testing. 

To assess performance, we benchmark \sathg\ against eighteen models using their default settings. These models are categorized based on their architectural design: models intended for standard graphs are referred to as graph-based models, while those designed for hypergraphs are referred to as hypergraph-based models. The baseline hypergraph-based models utilize hypergraphs constructed according to the methodologies described in their respective original studies. The graph-based models include Graph Transformer (GT) full and sparse versions \cite{dwivedi2020generalization}, Graphormer \cite{ying2021transformers}, ANS-GT \cite{zhang2022hierarchical}, NAGphormer \cite{chen2022nagphormer}, GPRGNN \cite{chien2021adaptive}, GCNII \cite{chen2020simple}. The hypergraph-based models include HGNN \cite{feng2019hypergraph}, HCHA \cite{bai2021hypergraph}, HyperGCN \cite{yadati2019hypergcn}, DHGNN \cite{jiang2019dynamic}, HNHN \cite{dong2020hnhn}, UniGCNII \cite{huang2021unignn}, AllSetTransformer \cite{chien2021you}, SheafHyperGCN and SheafHyperGNN \cite{duta2024sheaf}, HSL \cite{cai2022hypergraph}, and DHKH \cite{kang2022dynamic}. 

\begin{table}
 \centering
      \caption{Dataset Statistics. Here, \#N, \#E, and \#C are the number of nodes, number of edges, and number of classes, respectively. Furthermore, \#HyE* indicates the number of hyperedges in the hypergraphs derived from these datasets via our hypergraph construction method.}
      \normalsize
  \begin{tabular}{|c|c|c|c||c|}
    \hline
    \cellcolor{gray!60} \textbf{Dataset} &
     \cellcolor{gray!60} \textbf{\#N} &
     \cellcolor{gray!60} \textbf{\#E} &
         \cellcolor{gray!60} \textbf{\#C}  &
          \cellcolor{gray!60} \textbf{\#HyE*}
    \\
    \hline
   Cora & 2,708  & 5,429   & 7 & 263\\
    \hline
    Citeseer & 3,312& 4,715  &6 &563\\
    \hline
    PubMed & 19,717& 44,338   & 3 & 168\\
    \hline
     DBLP  & 17,716 & 105,734  & 4 & 739\\
    \hline
     LastFMAsia  &  7,624 & 55,612   & 18 & 46\\
    \hline
         Wiki  &  2,405 & 17,981   & 17 & 59\\
    \hline
  \end{tabular}
  \label{dataset_statistics}\vspace{-2mm} 
\end{table}

\begin{table*}[t!]
\addtolength{\tabcolsep}{1pt}
\centering
\caption{Performance Comparisons: Mean accuracy (\%) \text{$\pm$} standard deviation}\vspace{-2mm}
\normalsize
\begin{tabular}{|c|c|c|c|c|c|c|c|c|c|}
\hline
\cellcolor{gray!60}\textbf{Method}&
\cellcolor{gray!60} \textbf{Model} &
\cellcolor{gray!60} \textbf{Cora} &
\cellcolor{gray!60} \textbf{Citeseer} &
\cellcolor{gray!60} \textbf{PubMed} &
\cellcolor{gray!60} \textbf{DBLP} &
\cellcolor{gray!60} \textbf{LFMA} &
\cellcolor{gray!60} \textbf{Wiki} 
\\
\hline
\multirow{5}{*}{Graph-based}& GT-full & 63.10$\pm$0.52 &59.05$\pm$0.78 & 77.02$\pm$1.01 & 78.85$\pm$0.25 & 79.23$\pm$0.42&68.13$\pm$0.88\\
& GT-sparse & 72.04$\pm$0.30 &66.12$\pm$0.85 &80.11$\pm$0.78& 82.15$\pm$0.15 & 81.21$\pm$1.23&72.44$\pm$0.68\\
& Graphormer & 70.76$\pm$1.52 &67.52$\pm$0.56 & 82.17$\pm$0.56& 80.01$\pm$0.65 & 83.23$\pm$1.45&73.11$\pm$1.05\\
& ANS-GT & 86.70$\pm$0.82 &73.52$\pm$0.55 & 87.92$\pm$0.88& 82.52$\pm$1.65 & 86.88$\pm$0.63&76.24$\pm$1.08\\
& NAGphormer & 86.12$\pm$0.72 &71.55$\pm$1.15 & 87.51$\pm$0.38& 81.89$\pm$1.27 & 86.15$\pm$1.22 &76.55$\pm$0.75\\
& GPRGNN & 86.17$\pm$0.68 &76.11$\pm$0.22 & 86.96$\pm$0.68& 81.78$\pm$0.46 & 85.92$\pm$0.85&74.64$\pm$0.32\\
& GCNII &86.71$\pm$0.82&75.86$\pm$0.25&88.07$\pm$0.65&81.58$\pm$0.78& 85.96$\pm$0.60&74.66$\pm$0.58\\

\hline
\multirow{10}{*}{Hypergraph-based}&HGNN&79.39$\pm$1.36& 72.45$\pm$1.16&86.44$\pm$0.44&78.25$\pm$0.25&80.48$\pm$0.64&73.82$\pm$0.92\\
&HCHA&79.14$\pm$1.02&72.42$\pm$1.42&86.41$\pm$0.36&79.52 $\pm$0.80&82.44 $\pm$1.79&74.13$\pm$1.32\\
& HyperGCN &78.45$\pm$1.26&71.28$\pm$0.82&82.84$\pm$8.67&82.78 $\pm$1.23&80.20$\pm$1.56&74.22$\pm$1.34\\
& DHGNN&79.52$\pm$1.19&73.59$\pm$0.50&80.50$\pm$1.35&80.37 $\pm$0.62& 80.22$\pm$0.80&75.52$\pm$0.68\\
& HNHN&76.36$\pm$1.92&72.64$\pm$1.57&86.90$\pm$0.30&80.72$\pm$1.26& 84.17$\pm$1.15&75.14$\pm$0.92\\
& UniGCNII&78.81$\pm$1.05 &73.05$\pm$2.21 & 88.25$\pm$0.40&82.17 $\pm$0.36&84.49$\pm$1.54&77.18$\pm$1.22\\
& AllSetTransformer&78.59$\pm$1.47&73.08$\pm$1.20&88.72$\pm$0.37&83.15 $\pm$0.56&86.21$\pm$1.07&78.49$\pm$0.82\\
& SheafHyperGCN&80.06$\pm$1.12&73.27$\pm$0.50&87.09$\pm$0.71&82.37$\pm$ 0.12&86.88$\pm$0.32&77.42$\pm$0.72 \\
& SheafHyperGNN & 81.30$\pm$1.70&74.71$\pm$1.23&87.68$\pm$0.60&83.11 $\pm$0.23&87.14$\pm$0.55&78.17$\pm$0.92 \\

& HSL&79.88$\pm$0.73&73.79$\pm$1.62&-& - & - &-\\  
 & 
 DHKH&82.60$\pm$0.55&71.68$\pm$1.20&77.50$\pm$0.80&81.48$\pm$0.82&83.25$\pm$1.56&75.18$\pm$1.23\\
&\textbf{\sathg}&\textbf{88.48}$\pm$\textbf{1.10} & \textbf{78.09}$\pm$\textbf{0.6}&\textbf{89.01}$\pm$\textbf{0.3}& 
\textbf{84.38}$\pm$\textbf{1.2}& 
\textbf{88.58}$\pm$\textbf{1.35}&
\textbf{79.15}$\pm$\textbf{0.55}\\
\hline
\end{tabular}
\label{result} 
\end{table*}  

In the structural and spatial encoding methods described in Section \ref{structural_and_spatial}, we employ four different encoding functions to capture distinct aspects of the input graph and constructed hypergraph. To integrate local connectivity information with $lse$, we utilize a two-layer GCN implemented in Deep Graph Library (DGL) \cite{wang2019dgl}. To compute $ce$ and $ue$, we develop learnable encoding functions by utilizing PyTorch's learnable \textit{Embedding layer}.  
$pe$ is executed using PyTorch's learnable \textit{Linear layer} that leverages hypergraph Laplacian eigenvectors to generate positional embeddings for each node.

\begin{table}
\centering
\caption{Impact of local and global structural information on the model performance (accuracy)
.}\vspace{-4mm}
\normalsize
\begin{tabular}{|C{0.9cm}|C{0.60cm}|C{1.05cm}|C{1.1cm}|C{0.7cm}|C{0.8cm}|C{0.8cm}|}
\hline
\cellcolor{gray!60}\textbf{\sathg} &
\cellcolor{gray!60} \textbf{Cora} &
\cellcolor{gray!60} \textbf{Citeseer} &
\cellcolor{gray!60} \textbf{PubMed} &
\cellcolor{gray!60} \textbf{DBLP} &
\cellcolor{gray!60} \textbf{LFMA} &
\cellcolor{gray!60} \textbf{Wiki}
\\
\hline
LS & 83.12 & 73.89 & 85.32 & 81.13 & 83.86 & 75.23 \\
\hline 
GS & 79.68 & 70.56 & 82.88 & 79.32 & 80.36 & 72.22\\
\hline

\end{tabular}
\label{abalation1}\vspace{-3mm}
\end{table} 
In \sathg, we conduct semi-supervised node classification in a transductive setting, repeating experiments ten times with different random splits. Node and hyperedge features are one-hot encoded, and we employ a single-layer \sathg\ with Adam optimization. The optimal hyperparameters are determined through a grid search on the validation set. Based on our grid search, we chose a learning rate of 0.001 and a dropout rate of 0.5 for regularization. The LeakyReLU activation function is applied, and the model has four attention heads. Training spans 500 epochs, incorporating early stopping if the validation accuracy does not change for 100 consecutive epochs. The hidden dimension of \sathg\ is set to 64. \sathg\ is implemented using the DGL with PyTorch on a Tesla V100-SXM2-32GB GPU.


\subsection{Comparison with Baseline Methods}
We evaluate the performance of our model by conducting experiments on six distinct datasets and comparing the results with eighteen state-of-the-art baselines. The baselines are considered if their experimental results or codes are available. The outcomes, presented in Table \ref{result}, unambiguously demonstrate the superiority of our model across all datasets. Specifically, our model excels on the Cora dataset, achieving an impressive accuracy of $88.48\%$. This significantly surpasses the accuracy of the best-performing graph-based baseline model, GCNII, at $86.71\%$ and exceeds the top-reported accuracy of the hypergraph-based baseline, DHKH, which stands at $82.60\%$. In the case of the Citeseer dataset, our model attains an accuracy of $78.09\%$, outperforming the graph-based leading baseline GPRGNN with an accuracy of $76.11\%$, and the hypergraph-based top-performing baseline, SheafHyperGNN, at $74.71\%$. The trend continues with the PubMed, DBLP, LastFMAsia (LFMA), and Wiki datasets, where our model substantially outperforms the baselines. The results underscore our model's substantial enhancements in classifying the datasets, setting a new standard compared to existing state-of-the-art methods.

\begin{figure*}
    \centering
\includegraphics[width=\textwidth]{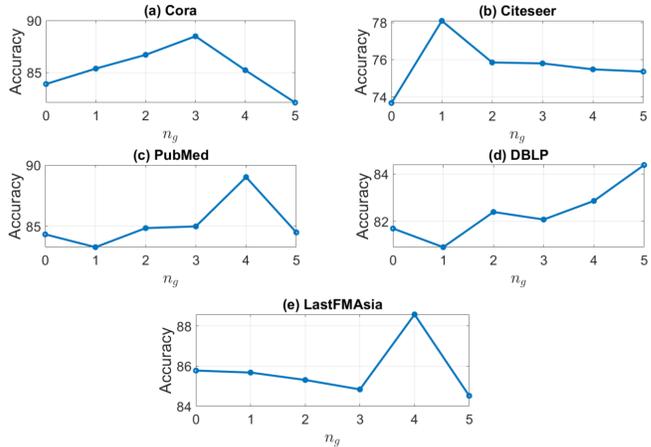}
   \caption{The performance (accuracy) of \sathg\ with different numbers of global nodes ($n_g$).}
    \label{fig:line_graph}
\end{figure*}

\begin{table}[t]
  \centering
  \caption{Impact of different design schemes on the model performance (accuracy) for Cora and PubMed datasets.}
  \label{abalation2}

  \setlength{\tabcolsep}{2pt} 
  \renewcommand{\arraystretch}{1.2} 
  \begin{tabularx}{\linewidth}{|c|*{8}{>{\centering\arraybackslash}X|}}
    \hline
    \cellcolor{gray!60}\diagbox[dir=NW,innerwidth=1.3cm]{\textbf{Dataset}}{\textbf{\sathg} \\ \scriptsize{W/O}} & \cellcolor{gray!60} \textbf{$pe$} & \cellcolor{gray!60} \textbf{$ce$} & \cellcolor{gray!60} \textbf{$ue$} & \cellcolor{gray!60} \textbf{$lse$} & \cellcolor{gray!60} \textbf{$lc$} & \cellcolor{gray!60} \textbf{$kc$} & \cellcolor{gray!60} \textbf{$hd$} & \cellcolor{gray!60} \textbf{$cc$} \\
    \hline
    Cora & 85.51 & 84.19 & 84.93 & 81.50 & 84.64 & 84.34 & 86.21 & 85.12 \\
    \hline
    PubMed & 86.22 & 85.52 & 86.68 & 82.44 & 85.88 & 86.72 & 87.34 & 86.57 \\
    \hline
  \end{tabularx}
\end{table}
A closer look at Table \ref{result} reveals that the performance of hyper-graph-based models is promising. This success might be attributed to their ability to efficiently learn the intricate higher-order structure within the hypergraph. In general, hypergraph-based baseline models present stiff competition, with DHGNN, AllSetTransformer, and SheafHyperGNN slightly outpacing others. Specifically, DHGNN, DHKH, and HSL acknowledge that the input hypergraph structure might not adequately represent the underlying relations in the data. Consequently, they simultaneously learn the hypergraph structure and hypergraph neural network, enabling them to prune noisy task-irrelevant and false-negative connections, producing better output. On the other hand, the AllSetTransformer framework blends Deep Sets \cite{zaheer2017deep} and Set Transformers \cite{lee2019set} with hypergraph neural networks to learn multiset functions. This synergy provides substantial modeling flexibility and expressive power, elevating the performance in various tasks. SheafHyperGNN and SheafHyperGCN enhance hypergraph representation by introducing cellular sheaves, a mathematical construction that adds additional structure to the conventional hypergraph while preserving their local, higher-order connectivity. This enhancement increases the expressivity of the models, enabling them to capture more nuanced and complex interactions within the hypergraph, leading to better overall representation.

Hypergraphs, known for their adeptness at representing intricate higher-order relationships, have become invaluable in numerous scientific endeavors. The proposed \sathg\ model builds on this, representing communities as hyperedges and thus forming a hypergraph that captures intricate node relations. In contrast to existing models that do not consider local connection information during hypergraph construction, \sathg\ leverages GCN to preserve this vital information. Additionally, \sathg\ integrates unique encoding modules to learn structural-spatial information and a structure-aware attention module. These modules enable \sathg\ to produce robust node representations, recognizing key nodes and hyperedges from both structural and semantic perspectives. 
\vspace{-1mm}
\subsection{Case Studies}
In this section, we present several case studies.\\
\textbf{ i. Impact of local/global structural knowledge} To delve deeper into the roles of local and global structural knowledge, we train \sathg\ excluding local structural (LS) components (specifically, $lse$, $ce$, $lc$, and $kc$) and omitting global structural (GS) elements (namely, $ue$, $hd$, and $cc$) on all dataset. Outcomes in Table \ref{abalation1} illustrate that local structural information significantly influences \sathg\ efficacy more than its global counterpart. Given that hypergraphs inherently harness higher-order global structural details from the graph, the omission of global encodings in \sathg\ does not critically hinder its performance. On the other hand, converting a graph to a hypergraph can result in a loss of local structural details, underscoring the importance of integrating local structural insights into hypergraph learning.
\\
\textbf{ii. Impact of different design schemes}
To better understand the impact of each design scheme, we conduct ablation studies on a small and a large dataset, Cora and PubMed, respectively. We consider running \sathg\ without (W/O) considering a specific design scheme each time. The results are presented in Table \ref{abalation2}, which implies the impact of the individual design scheme on the model performance. 
For instance, removing positional encoding ($pe$) results in an accuracy of 85.51\% for the Cora dataset and 86.22\% for the PubMed dataset. This table shows that for both datasets, the greatest performance degradation occurs when the local structure encoding ($lse$) is excluded from the input node feature. When we create hyperedge from a community, we lose the local connection information between the nodes, which might be very important for the hyperedges. $lse$ preserves local connection information by applying GCN to the input graph. The significant decrease in performance without $lse$ highlights its crucial contribution to the model's overall performance.
\\
\textbf{iii. Impact of Feature Normalization and Concatenation} 
As illustrated in Equation 4, our method combines various structural and spatial features (i.e., encodings) with the initial features of the nodes by summing them. To investigate the effect of normalizing these features before combining them with the node's initial features, we conduct an additional experiment. In this experiment, we normalize the features prior to their integration. The results of this experiment are presented in Table \ref{tab4} and are labeled as \sathg\textsubscript{N}. Additionally, we perform another experiment to evaluate the impact of concatenating the features instead of summing them. The outcomes of this approach are also shown in Table \ref{tab4} and are referred to as \sathg\textsubscript{C}. A comparative analysis of the different versions of \sathg\ presented in the table reveals that each version performs better on different datasets. However, the original \sathg\ method generally outperforms both the normalized variant (\sathg\textsubscript{N}) and the concatenated variant (\sathg\textsubscript{C}).
\\
\textbf{iv. Impact of global nodes} We investigate the effect of global nodes ($n_g$) on the model's performance, illustrated in Figure \ref{fig:line_graph}. As per Figure \ref{fig:line_graph} (a), the Cora dataset reveals a rising trend in accuracy as the global nodes $n_g$ increase, peaking at 88.48\% with $n_g=3$, followed by a decline. Analogously, Figure \ref{fig:line_graph} (b) presents the Citeseer dataset, where an increment in global nodes prompts an upsurge in accuracy, achieving a maximum of 78.09\% at $n_g=1$ before the accuracy trend reverses. Similarly, we get the best performance for the PubMed, DBLP, LastFMAsia, and Wiki datasets for $n_g=4$, $n_g=5$, $n_g=4$, and $n_g=4$ respectively. Overall, this trend suggests that an optimal number of global nodes can effectively incorporate relevant global information. However, exceeding this optimal number may introduce excessive parameters and reduced generalization ability. Thus, selecting an appropriate number of global nodes is crucial for optimal performance.

\begin{table}
\centering
\caption{Impact of feature normalization and concatenation on the model performance (accuracy).}\vspace{-2mm}
\normalsize
\begin{tabular}{|C{0.9cm}|C{0.60cm}|C{1.05cm}|C{1.1cm}|C{0.7cm}|C{0.8cm}|C{0.8cm}|}
\hline
\cellcolor{gray!60}\textbf{Model} &
\cellcolor{gray!60} \textbf{Cora} &
\cellcolor{gray!60} \textbf{Citeseer} &
\cellcolor{gray!60} \textbf{PubMed} &
\cellcolor{gray!60} \textbf{DBLP} &
\cellcolor{gray!60} \textbf{LFMA} &
\cellcolor{gray!60} \textbf{Wiki} 
\\
\hline
\sathg & 88.48 & \textbf{78.09} & \textbf{89.01} & 84.38 & \textbf{88.58} & \textbf{79.15}\\
\hline 
\sathg\textsubscript{N} & 86.88 & 76.23 & 88.41 & \textbf{85.46} & 87.26 & 78.13\\
\hline
\sathg\textsubscript{C} & \textbf{89.13} & 77.83 & 87.18 & 79.32 & 87.78 &75.25\\
\hline
\end{tabular}
\label{tab4}\vspace{-8mm}
\end{table}

\section{Conclusion}\label{sec:con}
In this paper, we introduce the Topology-guided Hypergraph Transformer Network (\sathg) as a novel hypergraph neural network model for node classification. While existing hypergraph transformer models mainly rely on semantic feature-based attention, they ignore topological information. The \sathg \ has addressed this limitation by presenting a structural and spatial encoding module along with a structure-aware attention mechanism. By considering both semantic and structural information, \sathg\ provides a comprehensive approach to node representation learning. Experimental findings underscore \sathg's superior node classification capabilities relative to benchmarks, highlighting its potential to revolutionize hypergraph analytics.


\bibliographystyle{ACM-Reference-Format}
\bibliography{main}


\begin{thebibliography}{39}


\ifx \showCODEN    \undefined \def \showCODEN     #1{\unskip}     \fi
\ifx \showDOI      \undefined \def \showDOI       #1{#1}\fi
\ifx \showISBNx    \undefined \def \showISBNx     #1{\unskip}     \fi
\ifx \showISBNxiii \undefined \def \showISBNxiii  #1{\unskip}     \fi
\ifx \showISSN     \undefined \def \showISSN      #1{\unskip}     \fi
\ifx \showLCCN     \undefined \def \showLCCN      #1{\unskip}     \fi
\ifx \shownote     \undefined \def \shownote      #1{#1}          \fi
\ifx \showarticletitle \undefined \def \showarticletitle #1{#1}   \fi
\ifx \showURL      \undefined \def \showURL       {\relax}        \fi
\providecommand\bibfield[2]{#2}
\providecommand\bibinfo[2]{#2}
\providecommand\natexlab[1]{#1}
\providecommand\showeprint[2][]{arXiv:#2}

\bibitem[Bai et~al\mbox{.}(2021)]%
        {bai2021hypergraph}
\bibfield{author}{\bibinfo{person}{Song Bai}, \bibinfo{person}{Feihu Zhang}, {and} \bibinfo{person}{Philip~HS Torr}.} \bibinfo{year}{2021}\natexlab{}.
\newblock \showarticletitle{Hypergraph convolution and hypergraph attention}.
\newblock \bibinfo{journal}{\emph{Pattern Recognition}}  \bibinfo{volume}{110} (\bibinfo{year}{2021}), \bibinfo{pages}{107637}.
\newblock


\bibitem[Cai et~al\mbox{.}(2022)]%
        {cai2022hypergraph}
\bibfield{author}{\bibinfo{person}{Derun Cai}, \bibinfo{person}{Moxian Song}, \bibinfo{person}{Chenxi Sun}, \bibinfo{person}{Baofeng Zhang}, \bibinfo{person}{Shenda Hong}, {and} \bibinfo{person}{Hongyan Li}.} \bibinfo{year}{2022}\natexlab{}.
\newblock \showarticletitle{Hypergraph structure learning for hypergraph neural networks}. In \bibinfo{booktitle}{\emph{Proceedings of the Thirty-First International Joint Conference on Artificial Intelligence, IJCAI-22, Lud De Raedt, Ed}}, Vol.~\bibinfo{volume}{7}. \bibinfo{pages}{1923--1929}.
\newblock


\bibitem[Chen et~al\mbox{.}(2020a)]%
        {chen2020hypergraph}
\bibfield{author}{\bibinfo{person}{Chaofan Chen}, \bibinfo{person}{Zelei Cheng}, \bibinfo{person}{Zuotian Li}, {and} \bibinfo{person}{Manyi Wang}.} \bibinfo{year}{2020}\natexlab{a}.
\newblock \showarticletitle{Hypergraph attention networks}. In \bibinfo{booktitle}{\emph{2020 IEEE 19th International Conference on Trust, Security and Privacy in Computing and Communications (TrustCom)}}. IEEE, \bibinfo{pages}{1560--1565}.
\newblock


\bibitem[Chen et~al\mbox{.}(2010)]%
        {chen2010detecting}
\bibfield{author}{\bibinfo{person}{Duanbing Chen}, \bibinfo{person}{Mingsheng Shang}, \bibinfo{person}{Zehua Lv}, {and} \bibinfo{person}{Yan Fu}.} \bibinfo{year}{2010}\natexlab{}.
\newblock \showarticletitle{Detecting overlapping communities of weighted networks via a local algorithm}.
\newblock \bibinfo{journal}{\emph{Physica A: Statistical Mechanics and its Applications}} \bibinfo{volume}{389}, \bibinfo{number}{19} (\bibinfo{year}{2010}), \bibinfo{pages}{4177--4187}.
\newblock


\bibitem[Chen et~al\mbox{.}(2022)]%
        {chen2022nagphormer}
\bibfield{author}{\bibinfo{person}{Jinsong Chen}, \bibinfo{person}{Kaiyuan Gao}, \bibinfo{person}{Gaichao Li}, {and} \bibinfo{person}{Kun He}.} \bibinfo{year}{2022}\natexlab{}.
\newblock \showarticletitle{NAGphormer: A tokenized graph transformer for node classification in large graphs}. In \bibinfo{booktitle}{\emph{The Eleventh International Conference on Learning Representations}}.
\newblock


\bibitem[Chen et~al\mbox{.}(2020b)]%
        {chen2020simple}
\bibfield{author}{\bibinfo{person}{Ming Chen}, \bibinfo{person}{Zhewei Wei}, \bibinfo{person}{Zengfeng Huang}, \bibinfo{person}{Bolin Ding}, {and} \bibinfo{person}{Yaliang Li}.} \bibinfo{year}{2020}\natexlab{b}.
\newblock \showarticletitle{Simple and deep graph convolutional networks}. In \bibinfo{booktitle}{\emph{International conference on machine learning}}. PMLR, \bibinfo{pages}{1725--1735}.
\newblock


\bibitem[Chien et~al\mbox{.}(2021a)]%
        {chien2021you}
\bibfield{author}{\bibinfo{person}{Eli Chien}, \bibinfo{person}{Chao Pan}, \bibinfo{person}{Jianhao Peng}, {and} \bibinfo{person}{Olgica Milenkovic}.} \bibinfo{year}{2021}\natexlab{a}.
\newblock \showarticletitle{You are allset: A multiset function framework for hypergraph neural networks}.
\newblock \bibinfo{journal}{\emph{arXiv preprint arXiv:2106.13264}} (\bibinfo{year}{2021}).
\newblock


\bibitem[Chien et~al\mbox{.}(2021b)]%
        {chien2021adaptive}
\bibfield{author}{\bibinfo{person}{Eli Chien}, \bibinfo{person}{Jianhao Peng}, \bibinfo{person}{Pan Li}, {and} \bibinfo{person}{Olgica Milenkovic}.} \bibinfo{year}{2021}\natexlab{b}.
\newblock \showarticletitle{Adaptive Universal Generalized PageRank Graph Neural Network}. In \bibinfo{booktitle}{\emph{International Conference on Learning Representations}}.
\newblock
\urldef\tempurl%
\url{https://openreview.net/forum?id=n6jl7fLxrP}
\showURL{%
\tempurl}


\bibitem[Ding et~al\mbox{.}(2023)]%
        {ding2023hyperformer}
\bibfield{author}{\bibinfo{person}{Kaize Ding}, \bibinfo{person}{Albert~Jiongqian Liang}, \bibinfo{person}{Bryan Perozzi}, \bibinfo{person}{Ting Chen}, \bibinfo{person}{Ruoxi Wang}, \bibinfo{person}{Lichan Hong}, \bibinfo{person}{Ed~H Chi}, \bibinfo{person}{Huan Liu}, {and} \bibinfo{person}{Derek~Zhiyuan Cheng}.} \bibinfo{year}{2023}\natexlab{}.
\newblock \showarticletitle{HyperFormer: Learning Expressive Sparse Feature Representations via Hypergraph Transformer}. In \bibinfo{booktitle}{\emph{Proceedings of the 46th International ACM SIGIR Conference on Research and Development in Information Retrieval}}. \bibinfo{pages}{2062--2066}.
\newblock


\bibitem[Dong et~al\mbox{.}(2020)]%
        {dong2020hnhn}
\bibfield{author}{\bibinfo{person}{Yihe Dong}, \bibinfo{person}{Will Sawin}, {and} \bibinfo{person}{Yoshua Bengio}.} \bibinfo{year}{2020}\natexlab{}.
\newblock \showarticletitle{Hnhn: Hypergraph networks with hyperedge neurons}.
\newblock \bibinfo{journal}{\emph{arXiv preprint arXiv:2006.12278}} (\bibinfo{year}{2020}).
\newblock


\bibitem[Duta et~al\mbox{.}(2024)]%
        {duta2024sheaf}
\bibfield{author}{\bibinfo{person}{Iulia Duta}, \bibinfo{person}{Giulia Cassar{\`a}}, \bibinfo{person}{Fabrizio Silvestri}, {and} \bibinfo{person}{Pietro Li{\`o}}.} \bibinfo{year}{2024}\natexlab{}.
\newblock \showarticletitle{Sheaf Hypergraph Networks}.
\newblock \bibinfo{journal}{\emph{Advances in Neural Information Processing Systems}}  \bibinfo{volume}{36} (\bibinfo{year}{2024}).
\newblock


\bibitem[Dwivedi and Bresson(2020)]%
        {dwivedi2020generalization}
\bibfield{author}{\bibinfo{person}{Vijay~Prakash Dwivedi} {and} \bibinfo{person}{Xavier Bresson}.} \bibinfo{year}{2020}\natexlab{}.
\newblock \showarticletitle{A generalization of transformer networks to graphs}.
\newblock \bibinfo{journal}{\emph{arXiv preprint arXiv:2012.09699}} (\bibinfo{year}{2020}).
\newblock


\bibitem[Feng et~al\mbox{.}(2019)]%
        {feng2019hypergraph}
\bibfield{author}{\bibinfo{person}{Yifan Feng}, \bibinfo{person}{Haoxuan You}, \bibinfo{person}{Zizhao Zhang}, \bibinfo{person}{Rongrong Ji}, {and} \bibinfo{person}{Yue Gao}.} \bibinfo{year}{2019}\natexlab{}.
\newblock \showarticletitle{Hypergraph neural networks}. In \bibinfo{booktitle}{\emph{Proceedings of the AAAI conference on artificial intelligence}}, Vol.~\bibinfo{volume}{33}. \bibinfo{pages}{3558--3565}.
\newblock


\bibitem[Fey and Lenssen(2019)]%
        {Fey/Lenssen/2019}
\bibfield{author}{\bibinfo{person}{Matthias Fey} {and} \bibinfo{person}{Jan~E. Lenssen}.} \bibinfo{year}{2019}\natexlab{}.
\newblock \showarticletitle{Fast Graph Representation Learning with PyTorch Geometric}. In \bibinfo{booktitle}{\emph{ICLR Workshop on Representation Learning on Graphs and Manifolds}}.
\newblock


\bibitem[Huang and Yang(2021)]%
        {huang2021unignn}
\bibfield{author}{\bibinfo{person}{Jing Huang} {and} \bibinfo{person}{Jie Yang}.} \bibinfo{year}{2021}\natexlab{}.
\newblock \showarticletitle{Unignn: a unified framework for graph and hypergraph neural networks}.
\newblock \bibinfo{journal}{\emph{arXiv preprint arXiv:2105.00956}} (\bibinfo{year}{2021}).
\newblock


\bibitem[Jiang et~al\mbox{.}(2019)]%
        {jiang2019dynamic}
\bibfield{author}{\bibinfo{person}{Jianwen Jiang}, \bibinfo{person}{Yuxuan Wei}, \bibinfo{person}{Yifan Feng}, \bibinfo{person}{Jingxuan Cao}, {and} \bibinfo{person}{Yue Gao}.} \bibinfo{year}{2019}\natexlab{}.
\newblock \showarticletitle{Dynamic Hypergraph Neural Networks.}. In \bibinfo{booktitle}{\emph{IJCAI}}. \bibinfo{pages}{2635--2641}.
\newblock


\bibitem[Kang et~al\mbox{.}(2022)]%
        {kang2022dynamic}
\bibfield{author}{\bibinfo{person}{Xiaojun Kang}, \bibinfo{person}{Xinchuan Li}, \bibinfo{person}{Hong Yao}, \bibinfo{person}{Dan Li}, \bibinfo{person}{Bo Jiang}, \bibinfo{person}{Xiaoyue Peng}, \bibinfo{person}{Tiejun Wu}, \bibinfo{person}{Shihua Qi}, {and} \bibinfo{person}{Lijun Dong}.} \bibinfo{year}{2022}\natexlab{}.
\newblock \showarticletitle{Dynamic hypergraph neural networks based on key hyperedges}.
\newblock \bibinfo{journal}{\emph{Information Sciences}}  \bibinfo{volume}{616} (\bibinfo{year}{2022}), \bibinfo{pages}{37--51}.
\newblock


\bibitem[Kreuzer et~al\mbox{.}(2021)]%
        {kreuzer2021rethinking}
\bibfield{author}{\bibinfo{person}{Devin Kreuzer}, \bibinfo{person}{Dominique Beaini}, \bibinfo{person}{Will Hamilton}, \bibinfo{person}{Vincent L{\'e}tourneau}, {and} \bibinfo{person}{Prudencio Tossou}.} \bibinfo{year}{2021}\natexlab{}.
\newblock \showarticletitle{Rethinking graph transformers with spectral attention}.
\newblock \bibinfo{journal}{\emph{Advances in Neural Information Processing Systems}}  \bibinfo{volume}{34} (\bibinfo{year}{2021}), \bibinfo{pages}{21618--21629}.
\newblock


\bibitem[Lee et~al\mbox{.}(2019)]%
        {lee2019set}
\bibfield{author}{\bibinfo{person}{Juho Lee}, \bibinfo{person}{Yoonho Lee}, \bibinfo{person}{Jungtaek Kim}, \bibinfo{person}{Adam Kosiorek}, \bibinfo{person}{Seungjin Choi}, {and} \bibinfo{person}{Yee~Whye Teh}.} \bibinfo{year}{2019}\natexlab{}.
\newblock \showarticletitle{Set transformer: A framework for attention-based permutation-invariant neural networks}. In \bibinfo{booktitle}{\emph{International conference on machine learning}}. PMLR, \bibinfo{pages}{3744--3753}.
\newblock


\bibitem[Li et~al\mbox{.}(2022)]%
        {li2022kpgt}
\bibfield{author}{\bibinfo{person}{Han Li}, \bibinfo{person}{Dan Zhao}, {and} \bibinfo{person}{Jianyang Zeng}.} \bibinfo{year}{2022}\natexlab{}.
\newblock \showarticletitle{KPGT: knowledge-guided pre-training of graph transformer for molecular property prediction}. In \bibinfo{booktitle}{\emph{Proceedings of the 28th ACM SIGKDD Conference on Knowledge Discovery and Data Mining}}. \bibinfo{pages}{857--867}.
\newblock


\bibitem[Li et~al\mbox{.}(2023)]%
        {li2023hypergraph}
\bibfield{author}{\bibinfo{person}{Mengran Li}, \bibinfo{person}{Yong Zhang}, \bibinfo{person}{Xiaoyong Li}, \bibinfo{person}{Yuchen Zhang}, {and} \bibinfo{person}{Baocai Yin}.} \bibinfo{year}{2023}\natexlab{}.
\newblock \showarticletitle{Hypergraph Transformer Neural Networks}.
\newblock \bibinfo{journal}{\emph{ACM Transactions on Knowledge Discovery from Data}} \bibinfo{volume}{17}, \bibinfo{number}{5} (\bibinfo{year}{2023}), \bibinfo{pages}{1--22}.
\newblock


\bibitem[Liu et~al\mbox{.}(2023)]%
        {liu2023meta}
\bibfield{author}{\bibinfo{person}{Jie Liu}, \bibinfo{person}{Lingyun Song}, \bibinfo{person}{Guangtao Wang}, {and} \bibinfo{person}{Xuequn Shang}.} \bibinfo{year}{2023}\natexlab{}.
\newblock \showarticletitle{Meta-HGT: Metapath-aware HyperGraph Transformer for heterogeneous information network embedding}.
\newblock \bibinfo{journal}{\emph{Neural Networks}}  \bibinfo{volume}{157} (\bibinfo{year}{2023}), \bibinfo{pages}{65--76}.
\newblock


\bibitem[Luo(2022)]%
        {luo2022shine}
\bibfield{author}{\bibinfo{person}{Yuan Luo}.} \bibinfo{year}{2022}\natexlab{}.
\newblock \showarticletitle{SHINE: SubHypergraph Inductive Neural nEtwork}.
\newblock \bibinfo{journal}{\emph{Advances in Neural Information Processing Systems}}  \bibinfo{volume}{35} (\bibinfo{year}{2022}), \bibinfo{pages}{18779--18792}.
\newblock


\bibitem[Rodriguez(2009)]%
        {rodriguez2009laplacian}
\bibfield{author}{\bibinfo{person}{JA Rodriguez}.} \bibinfo{year}{2009}\natexlab{}.
\newblock \showarticletitle{Laplacian eigenvalues and partition problems in hypergraphs}.
\newblock \bibinfo{journal}{\emph{Applied Mathematics Letters}} \bibinfo{volume}{22}, \bibinfo{number}{6} (\bibinfo{year}{2009}), \bibinfo{pages}{916--921}.
\newblock


\bibitem[Vaswani et~al\mbox{.}(2017)]%
        {vaswani2017attention}
\bibfield{author}{\bibinfo{person}{Ashish Vaswani}, \bibinfo{person}{Noam Shazeer}, \bibinfo{person}{Niki Parmar}, \bibinfo{person}{Jakob Uszkoreit}, \bibinfo{person}{Llion Jones}, \bibinfo{person}{Aidan~N Gomez}, \bibinfo{person}{{\L}ukasz Kaiser}, {and} \bibinfo{person}{Illia Polosukhin}.} \bibinfo{year}{2017}\natexlab{}.
\newblock \showarticletitle{Attention is all you need}.
\newblock \bibinfo{journal}{\emph{Advances in neural information processing systems}}  \bibinfo{volume}{30} (\bibinfo{year}{2017}).
\newblock


\bibitem[Veli{\v{c}}kovi{\'c} et~al\mbox{.}(2017)]%
        {velivckovic2017graph}
\bibfield{author}{\bibinfo{person}{Petar Veli{\v{c}}kovi{\'c}}, \bibinfo{person}{Guillem Cucurull}, \bibinfo{person}{Arantxa Casanova}, \bibinfo{person}{Adriana Romero}, \bibinfo{person}{Pietro Lio}, {and} \bibinfo{person}{Yoshua Bengio}.} \bibinfo{year}{2017}\natexlab{}.
\newblock \showarticletitle{Graph attention networks}.
\newblock \bibinfo{journal}{\emph{arXiv preprint arXiv:1710.10903}} (\bibinfo{year}{2017}).
\newblock


\bibitem[Wang et~al\mbox{.}(2019)]%
        {wang2019dgl}
\bibfield{author}{\bibinfo{person}{Minjie Wang}, \bibinfo{person}{Da Zheng}, \bibinfo{person}{Zihao Ye}, \bibinfo{person}{Quan Gan}, \bibinfo{person}{Mufei Li}, \bibinfo{person}{Xiang Song}, \bibinfo{person}{Jinjing Zhou}, \bibinfo{person}{Chao Ma}, \bibinfo{person}{Lingfan Yu}, \bibinfo{person}{Yu Gai}, \bibinfo{person}{Tianjun Xiao}, \bibinfo{person}{Tong He}, \bibinfo{person}{George Karypis}, \bibinfo{person}{Jinyang Li}, {and} \bibinfo{person}{Zheng Zhang}.} \bibinfo{year}{2019}\natexlab{}.
\newblock \showarticletitle{Deep Graph Library: A Graph-Centric, Highly-Performant Package for Graph Neural Networks}.
\newblock \bibinfo{journal}{\emph{arXiv preprint arXiv:1909.01315}} (\bibinfo{year}{2019}).
\newblock


\bibitem[Wu et~al\mbox{.}(2022)]%
        {wu2022nodeformer}
\bibfield{author}{\bibinfo{person}{Qitian Wu}, \bibinfo{person}{Wentao Zhao}, \bibinfo{person}{Zenan Li}, \bibinfo{person}{David~P Wipf}, {and} \bibinfo{person}{Junchi Yan}.} \bibinfo{year}{2022}\natexlab{}.
\newblock \showarticletitle{Nodeformer: A scalable graph structure learning transformer for node classification}.
\newblock \bibinfo{journal}{\emph{Advances in Neural Information Processing Systems}}  \bibinfo{volume}{35} (\bibinfo{year}{2022}), \bibinfo{pages}{27387--27401}.
\newblock


\bibitem[Xia et~al\mbox{.}(2022)]%
        {xia2022self}
\bibfield{author}{\bibinfo{person}{Lianghao Xia}, \bibinfo{person}{Chao Huang}, {and} \bibinfo{person}{Chuxu Zhang}.} \bibinfo{year}{2022}\natexlab{}.
\newblock \showarticletitle{Self-supervised hypergraph transformer for recommender systems}. In \bibinfo{booktitle}{\emph{Proceedings of the 28th ACM SIGKDD Conference on Knowledge Discovery and Data Mining}}. \bibinfo{pages}{2100--2109}.
\newblock


\bibitem[Yadati et~al\mbox{.}(2019)]%
        {yadati2019hypergcn}
\bibfield{author}{\bibinfo{person}{Naganand Yadati}, \bibinfo{person}{Madhav Nimishakavi}, \bibinfo{person}{Prateek Yadav}, \bibinfo{person}{Vikram Nitin}, \bibinfo{person}{Anand Louis}, {and} \bibinfo{person}{Partha Talukdar}.} \bibinfo{year}{2019}\natexlab{}.
\newblock \showarticletitle{Hypergcn: A new method for training graph convolutional networks on hypergraphs}.
\newblock \bibinfo{journal}{\emph{Advances in neural information processing systems}}  \bibinfo{volume}{32} (\bibinfo{year}{2019}).
\newblock


\bibitem[Ying et~al\mbox{.}(2021)]%
        {ying2021transformers}
\bibfield{author}{\bibinfo{person}{Chengxuan Ying}, \bibinfo{person}{Tianle Cai}, \bibinfo{person}{Shengjie Luo}, \bibinfo{person}{Shuxin Zheng}, \bibinfo{person}{Guolin Ke}, \bibinfo{person}{Di He}, \bibinfo{person}{Yanming Shen}, {and} \bibinfo{person}{Tie-Yan Liu}.} \bibinfo{year}{2021}\natexlab{}.
\newblock \showarticletitle{Do transformers really perform badly for graph representation?}
\newblock \bibinfo{journal}{\emph{Advances in Neural Information Processing Systems}}  \bibinfo{volume}{34} (\bibinfo{year}{2021}), \bibinfo{pages}{28877--28888}.
\newblock


\bibitem[Yun et~al\mbox{.}(2019)]%
        {yun2019graph}
\bibfield{author}{\bibinfo{person}{Seongjun Yun}, \bibinfo{person}{Minbyul Jeong}, \bibinfo{person}{Raehyun Kim}, \bibinfo{person}{Jaewoo Kang}, {and} \bibinfo{person}{Hyunwoo~J Kim}.} \bibinfo{year}{2019}\natexlab{}.
\newblock \showarticletitle{Graph transformer networks}.
\newblock \bibinfo{journal}{\emph{Advances in neural information processing systems}}  \bibinfo{volume}{32} (\bibinfo{year}{2019}).
\newblock


\bibitem[Zaheer et~al\mbox{.}(2017)]%
        {zaheer2017deep}
\bibfield{author}{\bibinfo{person}{Manzil Zaheer}, \bibinfo{person}{Satwik Kottur}, \bibinfo{person}{Siamak Ravanbakhsh}, \bibinfo{person}{Barnabas Poczos}, \bibinfo{person}{Russ~R Salakhutdinov}, {and} \bibinfo{person}{Alexander~J Smola}.} \bibinfo{year}{2017}\natexlab{}.
\newblock \showarticletitle{Deep sets}.
\newblock \bibinfo{journal}{\emph{Advances in neural information processing systems}}  \bibinfo{volume}{30} (\bibinfo{year}{2017}).
\newblock


\bibitem[Zhang et~al\mbox{.}(2022a)]%
        {zhang2022hegel}
\bibfield{author}{\bibinfo{person}{Haopeng Zhang}, \bibinfo{person}{Xiao Liu}, {and} \bibinfo{person}{Jiawei Zhang}.} \bibinfo{year}{2022}\natexlab{a}.
\newblock \showarticletitle{Hegel: Hypergraph transformer for long document summarization}.
\newblock \bibinfo{journal}{\emph{arXiv preprint arXiv:2210.04126}} (\bibinfo{year}{2022}).
\newblock


\bibitem[Zhang et~al\mbox{.}(2022b)]%
        {zhang-etal-2022-hegel}
\bibfield{author}{\bibinfo{person}{Haopeng Zhang}, \bibinfo{person}{Xiao Liu}, {and} \bibinfo{person}{Jiawei Zhang}.} \bibinfo{year}{2022}\natexlab{b}.
\newblock \showarticletitle{{HEGEL}: Hypergraph Transformer for Long Document Summarization}. In \bibinfo{booktitle}{\emph{Proceedings of the 2022 Conference on Empirical Methods in Natural Language Processing}}, \bibfield{editor}{\bibinfo{person}{Yoav Goldberg}, \bibinfo{person}{Zornitsa Kozareva}, {and} \bibinfo{person}{Yue Zhang}} (Eds.). \bibinfo{publisher}{Association for Computational Linguistics}, \bibinfo{address}{Abu Dhabi, United Arab Emirates}, \bibinfo{pages}{10167--10176}.
\newblock
\urldef\tempurl%
\url{https://doi.org/10.18653/v1/2022.emnlp-main.692}
\showDOI{\tempurl}


\bibitem[Zhang et~al\mbox{.}(2020)]%
        {zhang2020hypersagnn}
\bibfield{author}{\bibinfo{person}{Ruochi Zhang}, \bibinfo{person}{Yuesong Zou}, {and} \bibinfo{person}{Jian Ma}.} \bibinfo{year}{2020}\natexlab{}.
\newblock \showarticletitle{Hyper-{SAGNN}: a self-attention based graph neural network for hypergraphs}. In \bibinfo{booktitle}{\emph{International Conference on Learning Representations (ICLR)}}.
\newblock


\bibitem[Zhang et~al\mbox{.}(2022c)]%
        {zhang2022hierarchical}
\bibfield{author}{\bibinfo{person}{Zaixi Zhang}, \bibinfo{person}{Qi Liu}, \bibinfo{person}{Qingyong Hu}, {and} \bibinfo{person}{Chee-Kong Lee}.} \bibinfo{year}{2022}\natexlab{c}.
\newblock \showarticletitle{Hierarchical graph transformer with adaptive node sampling}.
\newblock \bibinfo{journal}{\emph{Advances in Neural Information Processing Systems}}  \bibinfo{volume}{35} (\bibinfo{year}{2022}), \bibinfo{pages}{21171--21183}.
\newblock


\bibitem[Zhao et~al\mbox{.}(2021)]%
        {zhao2021gophormer}
\bibfield{author}{\bibinfo{person}{Jianan Zhao}, \bibinfo{person}{Chaozhuo Li}, \bibinfo{person}{Qianlong Wen}, \bibinfo{person}{Yiqi Wang}, \bibinfo{person}{Yuming Liu}, \bibinfo{person}{Hao Sun}, \bibinfo{person}{Xing Xie}, {and} \bibinfo{person}{Yanfang Ye}.} \bibinfo{year}{2021}\natexlab{}.
\newblock \showarticletitle{Gophormer: Ego-Graph Transformer for Node Classification}.
\newblock \bibinfo{journal}{\emph{arXiv preprint arXiv:2110.13094}} (\bibinfo{year}{2021}).
\newblock


\bibitem[Zhu et~al\mbox{.}(2023)]%
        {zhu2023structural}
\bibfield{author}{\bibinfo{person}{Wenhao Zhu}, \bibinfo{person}{Tianyu Wen}, \bibinfo{person}{Guojie Song}, \bibinfo{person}{Liang Wang}, {and} \bibinfo{person}{Bo Zheng}.} \bibinfo{year}{2023}\natexlab{}.
\newblock \showarticletitle{On Structural Expressive Power of Graph Transformers}. In \bibinfo{booktitle}{\emph{Proceedings of the 29th ACM SIGKDD Conference on Knowledge Discovery and Data Mining}}. Association for Computing Machinery, \bibinfo{address}{New York, NY, USA}, \bibinfo{pages}{3628--3637}.
\newblock
\showISBNx{9798400701030}
\urldef\tempurl%
\url{https://doi.org/10.1145/3580305.3599451}
\showDOI{\tempurl}


\end{thebibliography}


\end{document}